\documentclass[runningheads]{llncs}
\usepackage{graphicx}
\usepackage{orcidlink}
\usepackage[sort&compress,numbers]{natbib}
\usepackage{multirow}
\usepackage{graphicx}
\usepackage{booktabs}

\begin{document}
\title{Low-Resourced Machine Translation for Senegalese Wolof Language}
\titlerunning{Wolof Low-Resourced MT}  
%

\author{Derguene Mbaye\inst{1,2}\orcidlink{0000-0002-7490-2731} \and Moussa Diallo\inst{1} \and Thierno Ibrahima Diop\inst{2}
}

\authorrunning{Derguene Mbaye et al.} 

\institute{Universit\'{e} Cheikh Anta Diop, Dakar, S\'{e}n\'{e}gal,\\
\email{derguene.mbaye@ucad.edu.sn}, \email{moussa.diallo@esp.sn} \\
\texttt{https://www.ucad.sn/}
\and
Baamtu, Dakar, S\'{e}n\'{e}gal \\
\email{thierno.diop@baamtu.com}}
\maketitle              
\begin{abstract}
Natural Language Processing (NLP) research has made great advancements in recent years with major breakthroughs that have established new benchmarks. However, these advances have mainly benefited a certain group of languages commonly referred to as resource-rich such as English and French. Majority of other languages with weaker resources are then left behind which is the case for most African languages including Wolof. In this work, we present a parallel Wolof/French corpus of 123,000 sentences on which we conducted experiments on machine translation models based on Recurrent Neural Networks (RNN) in different data configurations. We noted performance gains with the models trained on subworded data as well as those trained on the French-English language pair compared to those trained on the French-Wolof pair under the same experimental conditions.

\end{abstract}
\section{Introduction}
A Machine Translation (MT) system allows to switch from a textual sequence (or an audio source) in a source language, to the same sequence in the target language. For a long time, Statistical Machine Translation (SMT) systems \cite{koehn_2009} were the most popular approach before Neural Machine Translation (NMT) ones \cite{bahdanau} came along and achieved an increasingly higher performance. However, the quality of such systems has always been closely related to the amount of data used in their design \cite{koehn-knowles-2017-six}. Thus, state-of-the-art MT systems have been developed with sequence-to-sequence models using the attention mechanism \cite{luong-etal-2015-effective} as well as the Transformer architecture \cite{NIPS2017_3f5ee243}. Languages for which this binding does not represent a constraint, such as English, are said to be resource-rich and have several million sentence pairs; most other languages fall under the concept of "Low Resource" (LR).
However, the term "Low Resource" can encompass various aspects and can extend beyond the language to domains or tasks where little data is available even if the language is a resource-rich language. This is illustrated in \cite{nlplrlsurvey} where the concept of "Low Resource" is defined in three different aspects:  availability of task-specific labels, unlabeled language text, and auxiliary data. As shown in \cite{adebara-abdul-mageed-2022-towards}, most African languages fit into this description which makes the work of researchers difficult and contributes to the low representation of African languages in NLP research \cite{51306}. This is particularly the case for Wolof which, beyond the lack of data, is a language for which little work has been done in NLP.

An Automatic Speech Recognition (ASR) dataset on 4 african languages including Wolof was collected in \cite{gauthier-etal-2016-collecting} and used to design the first ASR system in this language. In \cite{nguer:hal-01311413}, the design of the first collaborative online dictionary in Wolof adapted to the LMF\footnote{Lexical Markup Framework} standard has been initiated. As part of the Dictionnaires Langue Africaine-Français (DiLAF)\footnote{\url{http://pagesperso.ls2n.fr/~enguehard-c/DiLAF/index.php}} project, researchers have produced several dictionaries on 7 African languages including Wolof. However, at the time of writing, all the dictionaries are available online except Wolof. The autors in \cite{dione-2012-morphological} explored the development of a finite-state based morphological analyzer for Wolof, the implementation and evaluation of an LFG-based parser for Wolof \cite{dione-2020-implementation} and the creation of a Universal Dependency (UD) treebank for Wolof \cite{dione-2019-developing} which is the first UD treebank within the Northern Atlantic branch of the Niger-Congo languages. In \cite{lo:hal-02054917}, the authors studied the design of a spellchecker for Wolof by presenting an approach based on a dictionary as a lexicon and a morphological analyzer of the Wolof language.

However, to the best of our knowledge, the only work exploring specifically Wolof French machine translation systems is that in \cite{nguer-etal-2020-sencorpus} where the authors presented a corpus of 70,000 Wolof French parallel sentences with which Word Embedding models as well as LSTM-based translation models were developed; and in \cite{dione-etal-2022-low} where the authors extended the corpus to 83,000 sentences with which they trained two neural machine translation systems for the French$\rightarrow$Wolof and Wolof$\rightarrow$French directions based on the Transformer architecture. However, the results presented in \cite{nguer-etal-2020-sencorpus} were reported in terms of accuracy making it difficult to evaluate the actual translation quality of their systems.
Multilingual neural machine translation systems including Wolof have also been developed such as in \cite{adelani-etal-2022-thousand} where authors leveraged existing pre-trained models to create low-resource translation systems for 16 African languages. The Meta's No Language Left Behind project\footnote{\url{https://ai.facebook.com/research/no-language-left-behind/}} which is capable of translating 200 languages between each other also includes the Wolof language.
Nevertheless, beyond Wolof, substantial work has been done on low-resource language NMT (LRL-NMT) in general.
The Masakhane community\footnote{\url{https://www.masakhane.io/}} proposes to address the challenge by targeting African languages with a participatory approach \cite{nekoto-etal-2020-participatory} including all relevant resource persons in the process leading to the production of MT datasets and benchmarks for over 30 languages. A detailed study of different approaches has been performed in \cite{Ranathunga2021NeuralMT} to address LRL-NMT and a set of guidelines has been defined to select the possible NMT techniques for a given LRL data setting. A set of experiments has been performed in \cite{icelandic-medium} on different translation systems, both neural and statistical based, to translate from English to Icelandic.
Most of these works, however, are based on the Transformer architecture, which is very data-intensive. Less recent architectures such as RNNs could perform better in low-resource environments because of the lower parameters required.

In this paper, we present a work in progress of French-Wolof parallel sentence data collection constituting to date, the largest corpus yet collected in this language pair with 123,000 sentences filtered and aligned at the sentence level. We then propose to go further regarding the work in \cite{nguer-etal-2020-sencorpus} and explore the performance of RNN models on our corpus by evaluating them with the BiLingual Evaluation Understudy (BLEU) metric \cite{papineni-etal-2002-bleu}, which is more representative than accuracy. Since subwording i.e. segmentation of the corpus into words or subwords, tends to improve the performance of translation models as shown in \cite{tok-impact}, we then experimented with the impact of this approach on our models. The paper is therefore organized as follows:
\begin{itemize}
  \item In Section \ref{wol-lang}, we present a describtion of the Wolof language.
  \item The data collection and filtering process are presented in Section \ref{data-filter}.
  \item Section \ref{exp} presents the experiments performed.
  \item The results are shown in Section \ref{results}.
  \item Section \ref{concl} concludes the work.
\end{itemize}

\section{The Wolof Language} \label{wol-lang}

As a West-Atlantic language mainly spoken in Senegal and Gambia, Wolof is also used in the Southern part of Mauritania. It belongs to the Atlantic group of the Niger-Congo language family and over seven million people spreading across three West African states is currently speaking Wolof. While only about 40\% of the Senegalese population are Wolof, about 90\% of the people speak the language as either their first, second or third language\footnote{\url{https://www.axl.cefan.ulaval.ca/afrique/senegal.htm}}.

There are two major geographical varieties of Wolof: one spoken in Senegal, and the other spoken in Gambia \cite{ethnologue-2019}. Even if people who speaks Wolof  understand  each  other,  the  Senegalese  Wolof  and the Gambian Wolof are two distincts languages:  both own their ISO 639-3 language code (respectively ”WOL” and ”WOF”). 
Although it has a long tradition of writing using the Arabic script known as Ajami or Wolofal, it has also been adapted to Roman script.

Wolof is an agglutinative language \cite{dione-2012-morphological} whose alphabet is quite close to the French one: we can find all the letters of its alphabet except H, V and Z \cite{adelani-etal-2021-masakhaner}. It also includes the characters $\eta$ ("ng") and $\widetilde{N}$ ("gn", as in Spanish). Accents are present, but in limited number ($\acute{A}$, $\acute{E}$, $\tilde{A}$, $\acute{O}$).
Twenty nine (29) Roman-based characters are used from the Latin script and most of them are involved in digraphs standing for geminate and prenasalized stops.  Unlike many other Niger-Congo languages, Wolof does not have tones. Nevertheless, Wolof syllables differ in intensity, e.g., long vowels are pronounced with more intensity than short ones. Length is represented by double vowel letters in writing and most Wolof consonants can be also geminated (doubled).
However, Wolof is not a standardized language (and some sources exclude the "H" from the alphabet) since no single variety has ever been accepted as the norm. Nonetheless, the Center of Applied Linguistics of Dakar (CLAD), coordinates the orthographic standardization of the Wolof language \cite{gauthier-etal-2016-collecting}.

\section{Data Collection} \label{data-filter}

\subsection{Corpus}
The construction of a dataset is a tedious and time-consuming task, especially for languages that have yet to be standardized like Wolof. The language is not taught in school and few people follow the spelling rules, which makes the texts available on sources such as social networks very heterogeneous and difficult to use. We therefore opted to collect data in French, since this is the official language in Senegal since colonization, and to have them translated by competent linguists to build part of the dataset from scratch. 

\begin{figure}[htbp]
\centerline{\includegraphics[width=1\textwidth, keepaspectratio]{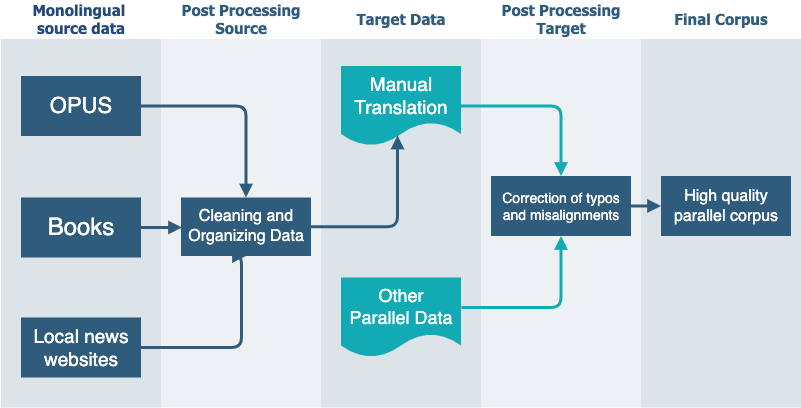}}
\caption{Data collection pipeline}
\label{fig}
\end{figure}

The linguists used the official Wolof alphabet established by the government\footnote{\url{http://www.jo.gouv.sn/spip.php?article4802}} to perform the translation. Monolingual french data are collected from existing resources such as Opus and text scraped from online sources that include news sites, religious and blogs. We used Opus to collect monolingual textual data in French and collected translations of the Quran and the Bible as parallel texts. We also collected data from offline sources such as French books that have been translated into Wolof. We were thus able to collect a corpus of 123,000 parallel French-Wolof sentences, making our corpus the largest collected to date.

For experimental purposes, the overall dataset is divided into three subsets: a training set, a validation set and a test set. The validation and test sets are kept fixed and separated from the full dataset with 16,000 sentences for the validation set and 7,000 for the test set. We only vary the training set from 10,000 to 100,000 sentences in steps of 10,000 sentences.

\subsection{Data filtering}
Before distributing the data between the different experimental configurations, we performed a set of post-processing operations. We started by performing stratified sampling to ensure that the validation and test sets were representative of the overall dataset and thus limit sampling bias.

Since the quality of the system depends directly on the quality of the data, we were inspired by the approaches proposed in \cite{pinnis-2018-tildes} to then filter our dataset. We have thus removed sentences written in the same language on both sides, duplicate pairs of sentences as well as sentences that are identical on both sides. We also removed special characters, URLs and filtered out sentences that were too long and under-represented in the dataset. We consider a sentence to be too long when its size (number of words) is greater than twice the average size of the sentences in the dataset considered.

\section{Experiments} \label{exp}
Despite having collected a corpus of 123,000 sentences, we are still in a low-resource configuration for the NMT. We have therefore opted for a medium data-intensive architecture (compared to SMT and Transformers) and exploited data manipulations to maximize the performance of the model.

We used OpenNMT \cite{klein-etal-2017-opennmt} to reproduce a similar architecture to that of \cite{nguer-etal-2020-sencorpus} in order to compare the results. The RNN model is thus composed of an LSTM layer \cite{lstm} at both the encoder and decoder with 300 hidden units and a dropout layer. The dropout rate is set to 0.1 and the embedding size to 128. We have defined an optimizer Adam \cite{Kingma2015AdamAM} with a learning rate of 0.001 and the batch size is set to 4096 tokens.

We split our dataset into different size configurations and in each configuration, the model is trained in the directions Fr$\rightarrow$Wo and Fr$\rightarrow$En until it reaches convergence. Convergence is considered to be reached when no improvement is observed on the validation set after 6 checkpoints.

For data subwording, we used SentencePiece \cite{kudo-richardson-2018-sentencepiece} with Byte-Pair Encoding (BPE) which offers interesting performance gains in agglutinative languages like Wolof \cite{sennrich-etal-2016-neural}. We then generated a vocabulary on all segments of the considered size configuration's training set and performed an automatic model evaluation using BLEU \cite{papineni-etal-2002-bleu}. BLEU is the most widely used metric in NMT in view of the fairly high correlation it has with human evaluations. We used the SacreBLEU \cite{post-2018-call} implementation\footnote{version 2.0.0} of the BLUE metric to evaluate the models.

\section{Results} \label{results}
\label{sec:bibtex}
We compare the same architectures in the same data size configurations (i) when the data are provided to the model in a raw form i.e. without subwording, compared to when they are subworded before training (ii) when they are trained on the different language pairs i.e. Fr$\rightarrow$En compared to Fr$\rightarrow$Wo.
The first case allows us to measure the impact of subwording on the quality of the translations and the second allows us to observe the influence of linguistic properties between languages that can facilitate or hinder translation performance.

Tables \ref{fr-wo-tab} and \ref{fr-en-tab} show the results of the translation experiments and all BLEU scores were computed on the test set. 


\begin{table}[htbp]
\caption{French Wolof Experimentation}
\begin{center}
\begin{tabular}{c c c}
\hline\rule{0pt}{12pt}
\textbf{Training size} & {\textbf{No subword}} & {\textbf{With subword}} \\[2pt]
\hline\rule{0pt}{12pt}
100k & 15.22 & 16.71 \\
90k &  14.41 & 15.28  \\
80k & 15.12 & 16.09 \\
70k & 12.76 & 14.85 \\
60k & 12.11 & 14.23 \\
50k & 10.45 & 12.14 \\
40k & 9.35 & 11.03 \\
30k & 7.33 & 9.73 \\
20k & 5.58 & 7.45 \\
10k & 3.94 & 4.84 \\[2pt]
\hline
\end{tabular}
\label{fr-wo-tab}
\end{center}
\end{table}

\begin{figure}[htbp]
\centerline{\includegraphics[width=0.7\textwidth, keepaspectratio]{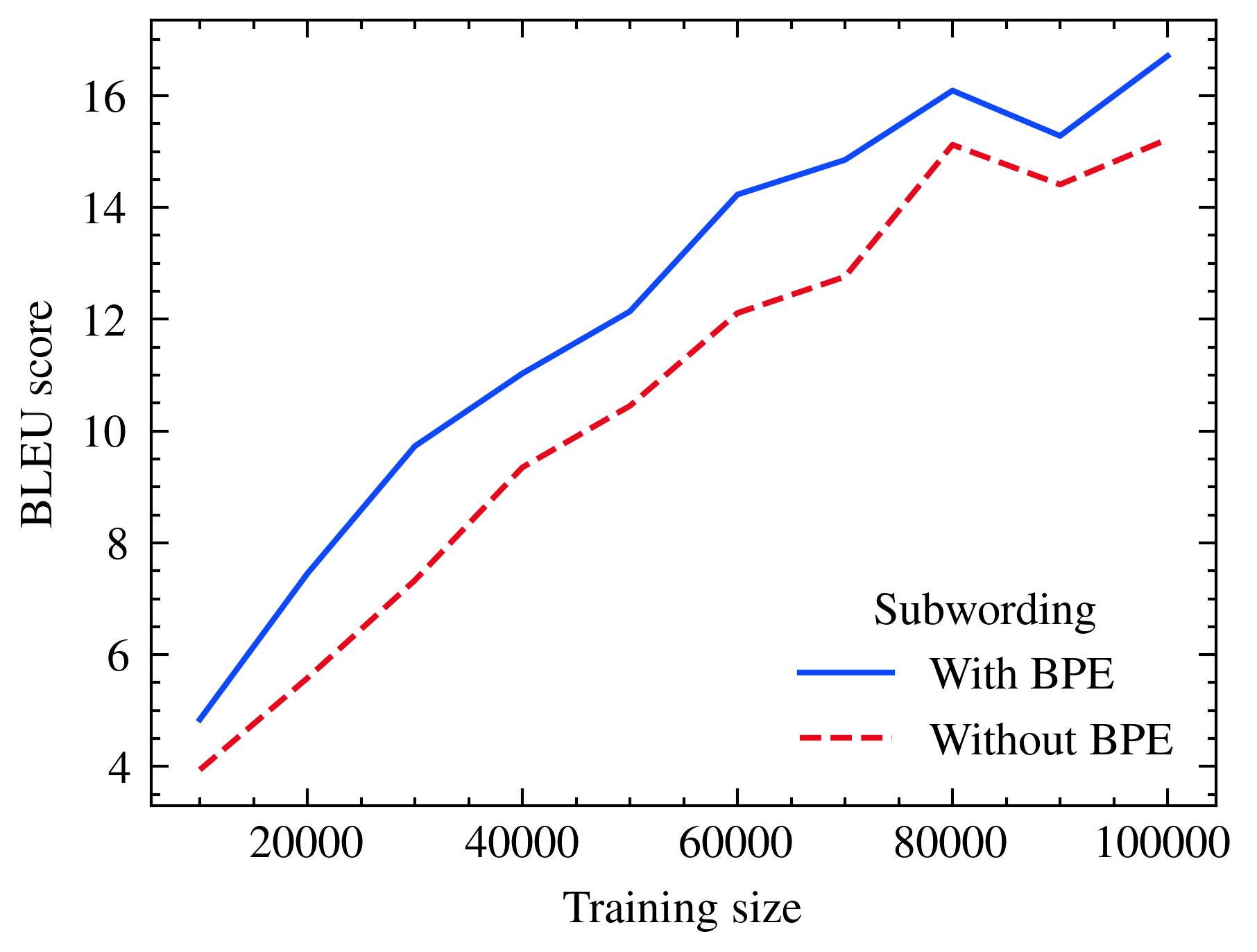}}
\caption{Performance evolution of Fr$\rightarrow$Wo NMT models on subworded and non-subworded data in the same data size configurations}
\label{fr-wo-fig}
\end{figure}

In Table~\ref{fr-wo-tab}, we observe a gain of about 1.6 point of BLEU score between the raw corpus and the subworded one on Fr$\rightarrow$Wo data, which can be explained by the fact that the subwords are more frequent and are therefore better learned by the model.
This gain is more visible with Fig.\ref{fr-wo-fig} where we see that the performance of the model on the subworded data is better at each training checkpoint.

\begin{table}[htbp]
\caption{French English Experimentation}
\begin{center}
\begin{tabular}{c c c}
\hline\rule{0pt}{12pt}
\textbf{Training size} & {\textbf{No subword}} & {\textbf{With subword}} \\[2pt]
\hline\rule{0pt}{12pt}

100k & 18.88 & 22.19 \\
90k & 18.52 & 21.11 \\
80k & 18.05 & 20.79 \\
70k & 17.82 & 20.57 \\
60k & 16.70 & 19.28 \\
50k & 15.17 & 18.94 \\
40k & 14.18 & 17.52 \\
30k & 4.68 & 16.22 \\
20k & 10.5 & 14.8 \\
10k & 3.34 & 10.1 \\[2pt]
\hline
\end{tabular}
\label{fr-en-tab}
\end{center}
\end{table}

We observe a similar pattern in Table~\ref{fr-en-tab} on Fr$\rightarrow$En data with a gain of about 4 points of BLUE score this time. When we compare the experimental results between the two language pairs, we also notice that under the same experimental conditions (corpus size and subwording), a gain of about 3.5 is noted on the BLEU score on Fr$\rightarrow$En data compared to Fr$\rightarrow$Wo.

\begin{figure}[htbp]
\centerline{\includegraphics[width=0.7\textwidth, keepaspectratio]{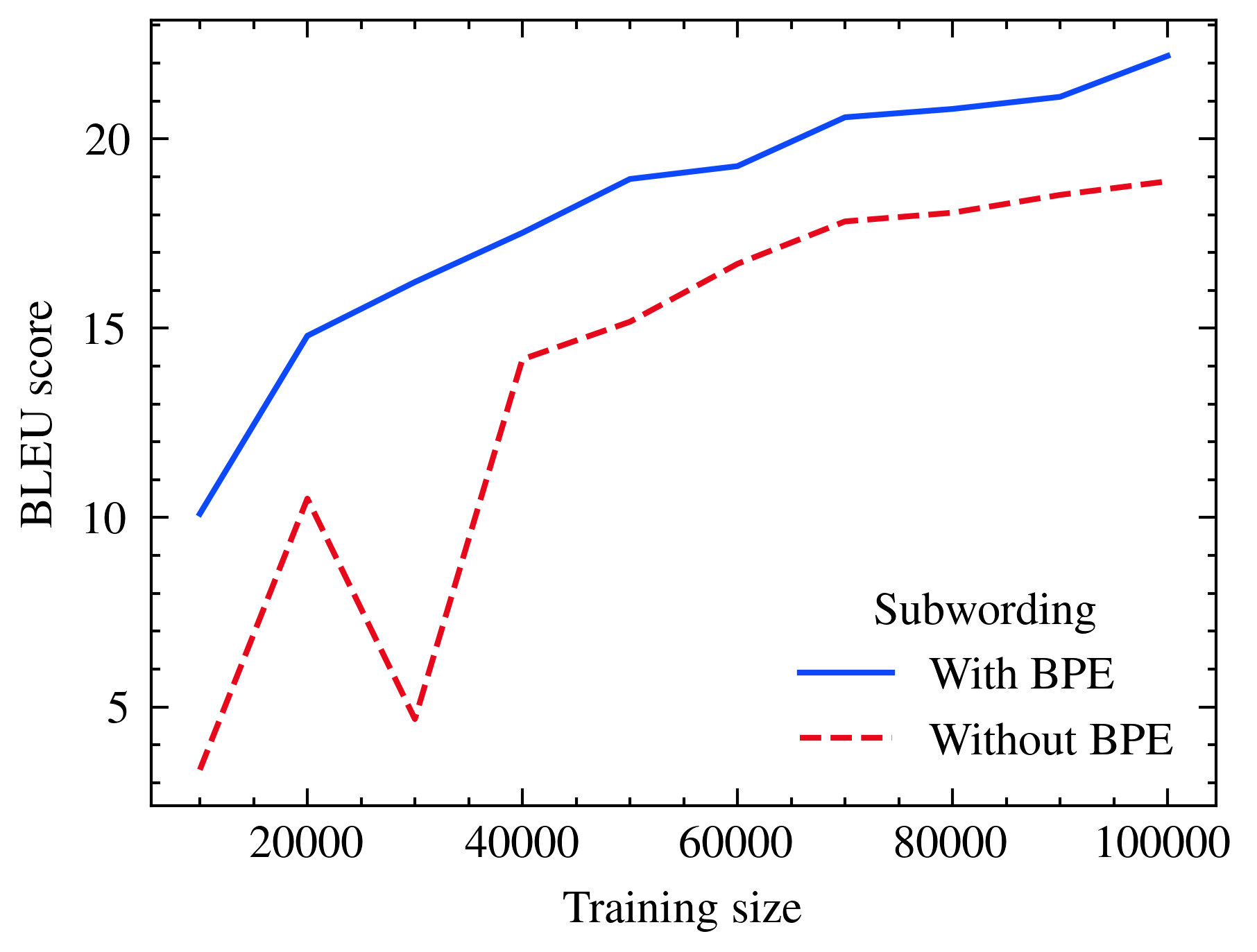}}
\caption{Performance evolution of Fr$\rightarrow$En NMT models on subworded and non-subworded data in the same data size configurations}
\label{fr-en-fig}
\end{figure}

Fig.\ref{fr-en-fig} illustrates well the behavior of the Fr$\rightarrow$En models on the different dataset formats with a sharp drop at checkpoint 30k. This is explained by the quality of the added data segment which contains a lot of artifacts and illustrates the fact that not all data points are useful for training.

\begin{figure}[htbp]
\centerline{\includegraphics[width=0.7\textwidth, keepaspectratio]{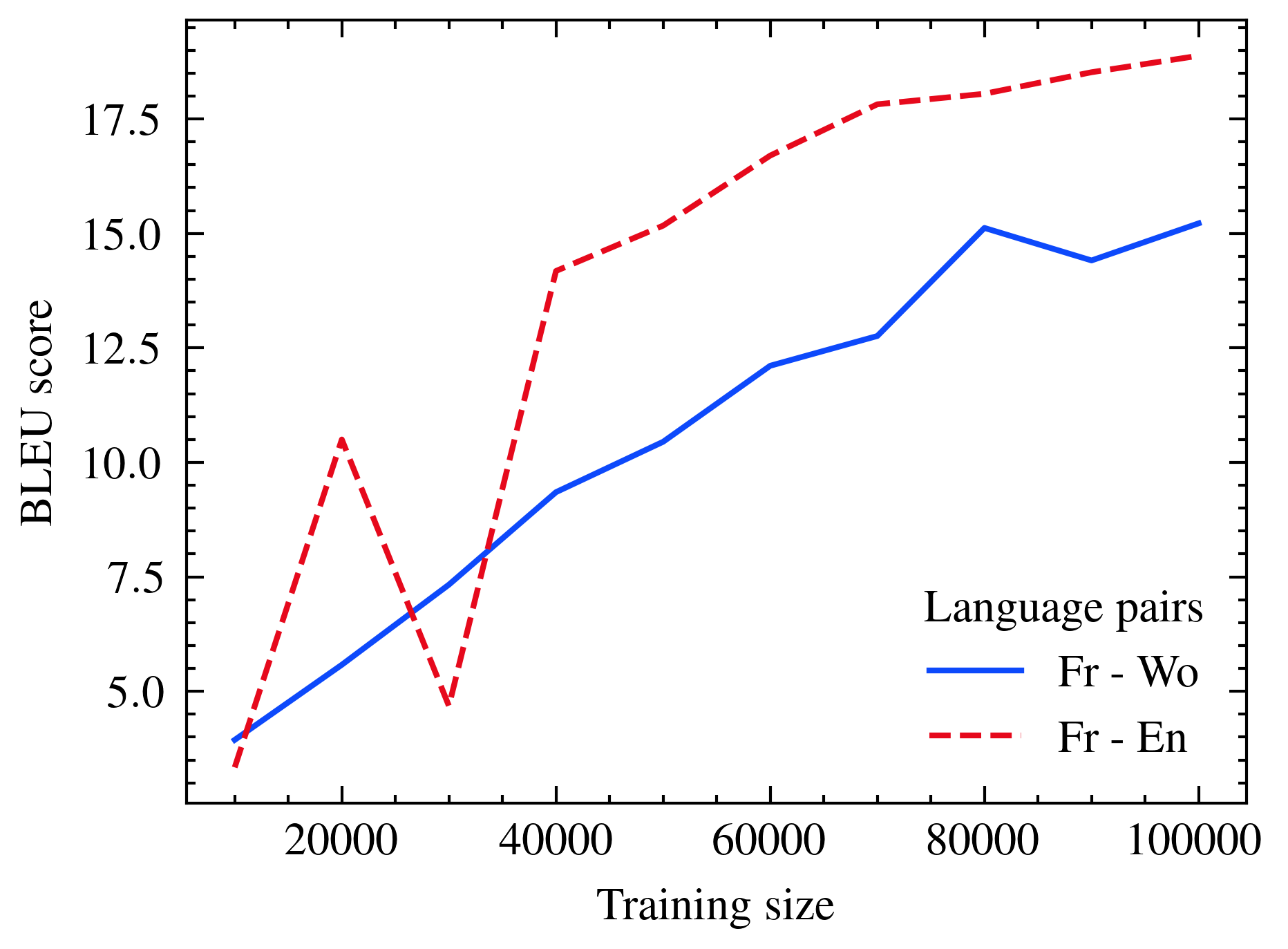}}
\caption{Performance evolution of Fr$\rightarrow$Wo and Fr$\rightarrow$En NMT models on raw data in the same data size configurations}
\label{nobpe-woen}
\end{figure}

\begin{figure}[htbp]
\centerline{\includegraphics[width=0.7\textwidth, keepaspectratio]{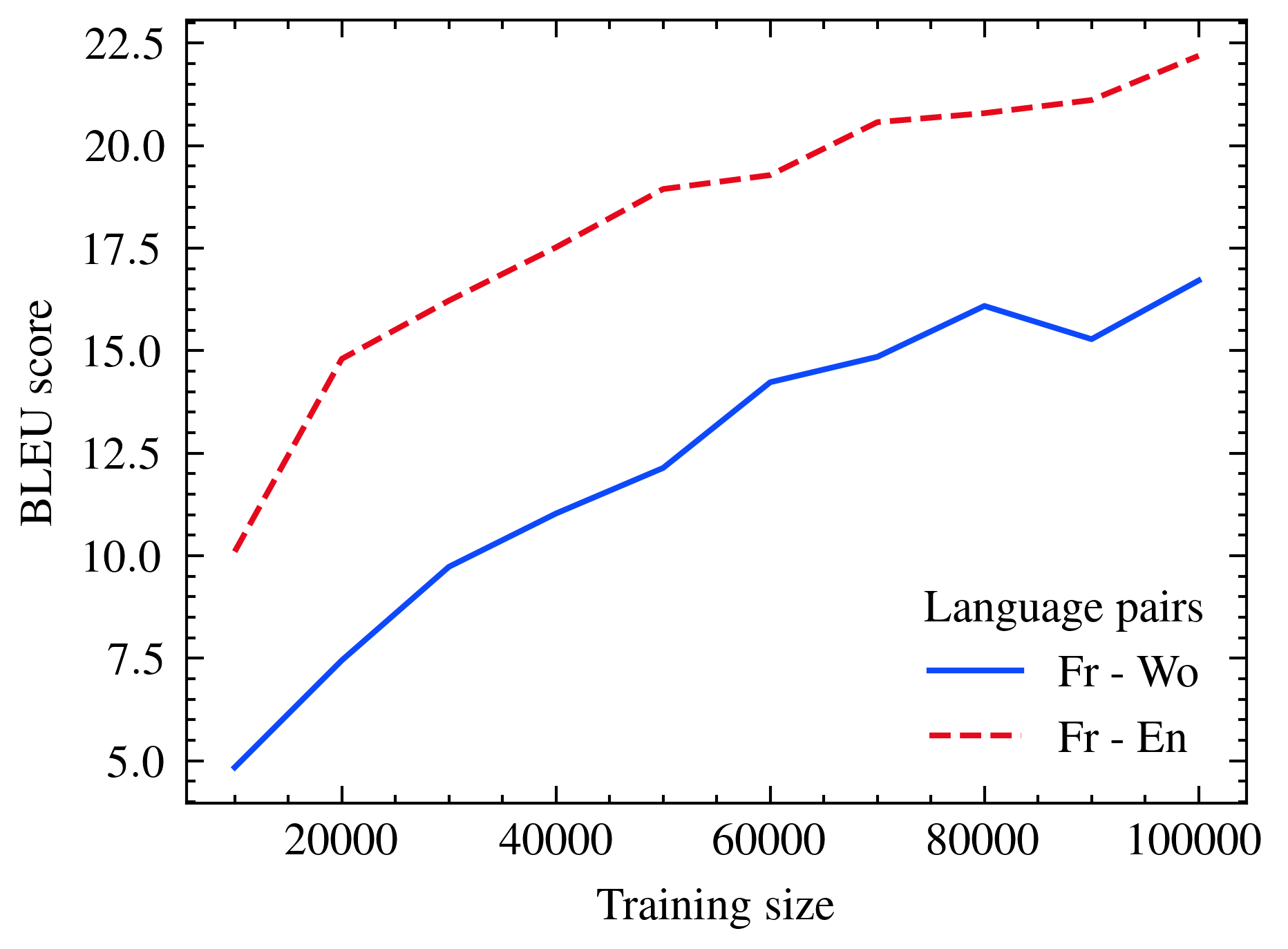}}
\caption{Performance evolution of Fr$\rightarrow$Wo and Fr$\rightarrow$En NMT models on subworded data in the same data size configurations}
\label{bpe-woen}
\end{figure}

In addition to subwording, we wanted to observe whether linguistic properties shared between two languages could influence translation performance. Fig.\ref{nobpe-woen} and Fig.\ref{bpe-woen} illustrate the performance of the two models in the same configurations (architecture and data) on the language pairs Fr$\rightarrow$Wo and Fr$\rightarrow$En.

In general, whether the data is subworded or not, we notice that the performance of the model trained on the Fr$\rightarrow$En language pair is better than the one trained on the Fr$\rightarrow$Wo language pair at all training checkpoints except the one at 30k where a sharp drop is observed. This can be explained by the linguistic similarities between French and English which, although belonging to different families, share the same alphabet. They also have a lexical similarity of 27\% \cite{ethnologue-2019} and words from one language that are found or have their origins in the other language. Our assumption is that the difference in morphology between the language pairs influences the ability of the model to translate one language into the other. The Wolof alphabet has more letters than the French one and Wolof is morphologically richer, which could hinder the ability of the model to capture the specificities of this language.

\section{Conclusion} \label{concl}

In this article, we presented a French Wolof parallel corpus of 123,000 sentences. This corpus was mostly collected from scratch, as openly accessible resources concerning this pair are scarce. As the collection project is still in progress, the dataset is not yet open. We then conducted experiments on various architectures of LSTM and global attention based neural machine translation models and showed that these systems were more efficient on subworded data. Further experiments attempted to investigate the impact of linguistic similarity between a language pair on translation performance by comparing systems on two different language pairs under the same experimental conditions: Fr$\rightarrow$Wo and Fr$\rightarrow$En. 

To the best of our knowledge, our corpus constitutes the largest corpus yet collected in this language pair and it is the first work where LSTM-based machine translation systems specifically for the Fr$\leftrightarrow$Wo language pair are presenting the performance with the BLEU metric which allows to better appreciate the performance of NMT models.

However, the BLUE metric may induce biases and therefore not be sufficient for a complete evaluation of the actual quality of our systems \cite{wieting-etal-2019-beyond}. Subwording also brought significant gains, but the SentencePiece method is language agnostic and may not be optimal for all languages.
On the other hand, RNN systems suffer from the inability to handle long sequences even when LSTM or GRU \cite{lstm} cells are used. State of the art systems today are mainly based on the Transformer architecture which has a better ability to handle longer sequences and allows parallelization as it does not do sequential processing. Cross-lingual transfer learning approaches have also shown very promising results in addressing machine translation for low-resource languages and are thus a relevant direction to explore.

In future work, we plan to further extend our dataset and explore Transformer-based models that, although data-intensive, can be optimized for a limited resource configuration \cite{araabi-monz-2020-optimizing}. We will also do a comparative analysis of multilingual models in order to choose the one that has better transfer learning performance with Wolof and perform transfer learning on it.

%
%
\bibliographystyle{spbasic_unsrt}
\bibliography{mybibliography}

\end{document}